%% file: main.tex
\documentclass[review]{elsarticle}

\usepackage{hyperref}
\usepackage{lineno}
\modulolinenumbers[5]

\usepackage[english]{babel}		% Idioma del documento
\usepackage[utf8]{inputenc}
\usepackage{color}
\usepackage{braket}				% Notacion braket
\usepackage{amsfonts}
\usepackage{amsmath}
\usepackage{amsthm}
\usepackage{graphicx}
\usepackage{geometry}				% Todos los parametros geometricos de la hoja
\usepackage[shortlabels]{enumitem}
\usepackage{caption}
\usepackage{subcaption}
\usepackage{float}  				% me deja poner el grafico [H]
\usepackage{mathtools} 			% align
\usepackage{import}
\usepackage{lmodern}				% que hace esto?
\usepackage[toc,page]{appendix} % Incluir apendices de manera 
\usepackage{cleveref}
\usepackage[margin=20pt,font={small,it},labelfont=bf]{caption} % captions 
\usepackage{epigraph}				% para poner quotes 

\usepackage{listings}%para poner codigo directamente y en formato re cool
\usepackage{natbib}
\setcitestyle{authoryear,open={[},close={]}}

\graphicspath{ {images/} }      % Carpeta de imagenes

\definecolor{warning}{rgb}{1,0,0}

\definecolor{naranja}{rgb}{1,0.5,0}
\definecolor{blue}{rgb}{0,0,1}
\definecolor{green}{rgb}{0,0.5,0}
\definecolor{red}{rgb}{0.8,0,0}
\definecolor{violeta}{rgb}{0.5,0,0.5}
\definecolor{ro}{rgb}{0.5,0,0}
\definecolor{codegreen}{rgb}{0,0.6,0}
\definecolor{codegray}{rgb}{0.5,0.5,0.5}
\definecolor{codepurple}{rgb}{0.58,0,0.82}
\definecolor{backcolour}{RGB}{200,200,210}

\definecolor{dkgreen}{rgb}{0,0.6,0}
\definecolor{gray}{rgb}{0.5,0.5,0.5}
\definecolor{mauve}{rgb}{0.58,0,0.82}

\date{}

\journal{Arxiv}

\begin{document}

% example: https://www.elsevier.com/authors/author-schemas/latex-instructions\textbf
\begin{frontmatter}
\title{Uncovering differential equations from data with hidden variables}

%% Group authors per affiliation:
\author{Agustín Somacal \fnref{aristas}}
\ead{a.somacal@aristas.com.ar}
% \address{}
\fntext[aristas]{Aristas S.R.L., Dorrego 1940, Torre A, 2do Piso, dpto. N (1425), Ciudad Autónoma de Buenos Aires, Argentina}

\author{Yamila Barrera\fnref{aristas}}
\ead{y.barrera@aristas.com.ar}
\address{}
% \fntext[aristas]{Aristas S.R.L., Dorrego 1940, Torre A, 2do Piso, dpto. N (1425), Ciudad Autónoma de Buenos Aires, Argentina}

\author{Leonardo Boechi\fnref{IC}}
\ead{lboechi@ic.fcen.uba.ar}

\author{Matthieu Jonckheere\fnref{IC}}
\ead{mjonckhe@dm.uba.ar}
\fntext[IC]{Instituto de Calculo-CONICET, Intendente Güiraldes 2160, Ciudad Universitaria, Pabellón II, 2do. piso, (C1428EGA), Buenos Aires, Argentina}

\author{Vincent Lefieux\fnref{RTE}}
\ead{vincent.lefieux@rte-france.com}
\fntext[RTE]{Réseau de Transport d'Electricité (RTE), France.}

\author{Dominique Picard\fnref{ParisVII}}
\ead{picard@math.univ-paris-diderot.fr}

\fntext[ParisVII]{Université de Paris, LPSM, UFR Mathematiques Batiment Sophie Germain, 75013 Paris.}

\author{Ezequiel Smucler\fnref{diTella, aristas}}
\ead{e.smucler@aristas.com.ar}
\fntext[diTella]{Universidad Torcuato Di Tella,  Av. Figueroa Alcorta 7350 (C1428BCW) Sáenz Valiente 1010 (C1428BIJ) Ciudad de Buenos Aires, Argentina}
% \tnotetext[mytitlenote]{Github on \href{http://agusomacal}

% \tableofcontents

\begin{abstract}
 \input{sections/Abstract}
\end{abstract}

\begin{keyword}
\texttt{Differential equations} \sep \texttt{Dynamical systems}\sep  \texttt{Lasso} \sep  \texttt{Latent variables} \sep \texttt{
Machine learning}
% \MSC[2010] 00-01\sep  99-00
\end{keyword}

\end{frontmatter}

% \linenumbers

\section{Introduction}
\label{sec:intro}

\input{sections/Introduction}

\section{Methods}
\label{sec:methods}
\input{sections/Methodology}

\section{Results}
\label{sec:results}
\input{sections/Results}

\section{Discussion}
\label{sec:discussion}
\input{sections/Conclusions}

\section{Acknowledgements}

Funding: This work was supported by Aristas S. R. L. and RTE (Réseau de Transport d'Électricité).

%%%\nocite{}

% \cleardoublepage

\bibliography{main}
%\printbibliography

\end{document}

%% file: sections/Abstract.tex
SINDy is a method for learning system of differential equations from data by solving a sparse linear regression optimization problem  \citep{DataDrivenPDE2}. In this article, we propose an extension of the SINDy method that learns systems of differential equations in cases where some of the variables are not observed. Our extension is based on regressing a higher order time derivative of a target variable onto a dictionary of functions that includes lower order time derivatives of the target variable. We evaluate our method by measuring the prediction accuracy of the learned dynamical systems on synthetic data and on a real data-set of temperature time series provided by the Réseau de Transport d'Électricité (RTE). Our method provides high quality short-term forecasts and it is orders of magnitude faster than competing methods for learning differential equations with latent variables.

%% file: sections/Introduction.tex
Many branches of science are based on the study of dynamical systems. Examples include meteorology, biology and physics. The usual way to model deterministic dynamical systems is by using (partial) differential equations. Typically, differential equations models for a given dynamical system are derived using apriori insights into the problem at hand; then the model is validated using empirical observations. In an era in which massive data-sets pertaining to different fields of science are widely available, an interesting problem is whether it is possible for a useful differential equations model to be learned directly from data, without any major modeling effort required by the researcher.

The SINDy method is an approach of sparse identification of nonlinear dynamical systems that consists of linking the dynamical system discovery problem to a statistical regression problem (\citep{DataDrivenPDE2}, \citep{DataDrivenPDE} and \citep{sindy}). The main idea of the SINDy method is to consider a set of differential operators (possibly in both time and space if appropriate), discretize them, for example by using finite differences, and then regress the outcome of interest on the discretized differential operators. By solving the regression problem using an ad-hoc thresholded least-squares algorithm, they can build sparse, interpretable models, that use mostly low order derivatives. The authors explored the applicability of their method on simulated data, but only in situations in which all the variables of the simulated models are observed. %We provide further details of their approach in Section {\ref{sec:methods}}. 
%\agus{we say this also afterward} We highlight that in place of the thresholded least-squares algorithm, any other regularised linear regression estimator could be used. 

Our goal is to extend the SINDy model for the case in which not all relevant variables are observed, that is, in cases in which the main variable of interest depends on other variables of which no measurements are available. As an example of application we consider the climate time series of the Réseau de Transport d'Électricité (RTE). RTE is the main electricity network operational manager in France, who is interested in understanding the behavior of climate time series because of their impact on energy consumption. RTE uses high-level simulations of hourly temperature series to study the impact different climate scenarios have on electricity consumption, and hence on the French electrical power grid. The underlying simulations are based on the Navier-Stokes equations and include  variables as wind velocity, density, pressure, etc. The resulting dynamic system is known to be chaotic, see \citep{ChaosNavierStokes}. For that reason, our goal is to learn a system of differential equations that adequately models the dynamics of the temperature time series if the only observable variable is temperature, that is, if pressure, wind velocity, etc, are hidden variables.

To accommodate the possibility of hidden variables we note that, for a large class of dynamical systems, it is possible to reconstruct a trajectory (equivalent to the original one) given only one of the model variables, using its higher order derivatives (\citep{Takens}).
%Based on this result, we propose to augment the methodology developed in \citep{DataDrivenPDE2}, \citep{DataDrivenPDE} and \citep{sindy} by regressing higher-order time derivatives, to tackle situations in which not all relevant variables are observed.
%We estimate the coefficients of the dynamical system using the Lasso estimator: an $\ell_{1}$-regularized least squares regression estimator (\citep{lasso}). We choose to use the Lasso due to its simplicity, the abundance of theoretical guarantees on its performance (\citep{sparse-book}) and the availability of efficient algorithms to solve the convex optimization problem that defines the estimator.
%After estimating the regression coefficients, we build a forecasting method by integrating the retrieved differential equation.
%We call this methodology Latent ODE find (L-ODEfind).

Related to our approach is the Generalized Polynomial Modeling method (known as GPoMo), that addresses the recovery problem via a combinatorial search among a predefined set of polynomial functions of the observable variables  \citep{GPoMo} and \citep{Bongard9943}. The GPoMo method proceeds by choosing iteratively a family of combination of terms that minimize the Akaike or the Bayesian information criterion. Finally, it returns the set of best models. The authors also discuss the ability of their algorithm to find equations able to capture the dynamics when only some variables are observed. However, we will show that unfortunately this approach does not scale to large problems. Other approaches for learning dynamical systems from data available in the literature, such as those based on symbolic regression (\citep{symbolic1}), also have the drawback of being too computationally expensive.

%Evaluating the equations found by these methods can be challenging. When there are hidden variables, we propose to evaluate the solutions found by a given method for learning differential equations by measuring their forecasting accuracy. That is, the differential equations found by a method are integrated to obtain predictions which can then be compared with the true values. 

The article is organized as follows. In Section \ref{sec:methods} we review the GPoMo and SINDy methods and afterwards,  describe our methodology (named L-ODEfind) in detail. 
Section \ref{sec:results} presents the results of our experiments. 
In particular, in Section \ref{sec:fullinfo}, we compare the performance of GPoMo and L-ODEfind in recovering differential equations using empirical data in the case in which all relevant variables are observed. 
In Section \ref{sec:latent} we compare them in the harder case in which at least one relevant variable driving the dynamical system is latent. 
We apply our proposed method to real world temperature times series in Section \ref{sec:rte}. 
Finally, in Section \ref{sec:discussion} we discuss future work and possible extensions.

%% file: sections/Methodology.tex
\subsection{GPoMo: Differential equations recovery as a combinatorial search problem}

GPoMo is a method proposed and implemented by \citep{GPoMo} that addresses the differential equations recovery problem via a combinatorial search in the space of differential equations that can be expressed as polynomial functions of the observed variables. The method uses a genetic algorithm in which at each step new test models are generated by randomly choosing some polynomial terms to be included in the equations. This choice is made by making small variations (take or add a few terms) over the best previously seen models. Then, to select the winning models at each step, they are integrated and compared to the original data.

This combination of combinatorial search and integration steps makes the method slow (as we will see in the experiments). Moreover, except for polynomial combinations of the variables, it does not allow other types of regressors to be included, such as functions of the time variable. 
To overcome the aforementioned limitations of GPoMo, we based our approach on SINDy, which uses sparse regression to discover governing physical equations from measurement data. % For this method to work, on the one hand, the target variables must be specified. They will be a function of the data and, in particular, as we want to find a differential equation model, they will be some (discrete) time derivative of the observed variables. On the other hand, a set of regressors need to be build out of the domain (time, space) and observed variables to complete the regression model. 
We briefly review the SINDy method in the following section.

\subsection{SINDy: Differential equations recovery as a linear regression problem}

In our work we focus on ordinary differential equations even though SINDy can also tackle partial differential equations. In particular, consider a dynamical system represented by functions $f_1(t) \dots f_H(t)$ satisfying a set of differential equations of the form
\begin{equation}
    \mathbb{D}\mathbf{f} = \mathbf{U}(\mathbf{f}, \mathbb{E}\mathbf{f}),
    \label{eq:dyn_sys}
\end{equation}
where $\mathbf{f}=(f_1,\dots, f_H)$, $\mathbb{D},\mathbb{E}$ are differential operators in the temporal variables ($t$) and $\mathbf{U}: \mathbb{R}^{J}\to\mathbb{R}^{H}$ is an unknown map.
Suppose we have a series of $T$ equally spaced in time measurements, that is, we observe 
$
    f_h(t_i) \quad i \in \{1, \dots, T\}, h \in \{1, \dots, H\}.
$
An example of a dynamical system we will study in this paper is the classical Rossler system \eqref{eq:Rosseler} (\citep{Rosseler}). This system was originally designed to have similar properties and be simpler than the Lorenz system (\citep{lorenz1963}) which was, at the same time, a simplified model for atmospheric convection. The system is given by
 \begin{align}
 \frac{d f_{1}(t)}{dt} &= -f_{2}(t)-f_{3}(t), \nonumber \\
 \frac{d f_{2}(t)}{dt} &= f_{1}(t) + \alpha f_{2}(t) 
 \label{eq:Rosseler}, \\
 \frac{d f_{3}(t)}{dt} &= \beta + f_{3}(f_{1}(t) - \gamma), \nonumber
 \end{align}
 for constants $\alpha,\beta,\gamma$. This system can be written in the form \eqref{eq:dyn_sys} by taking $\mathbf{f}=(f_{1},f_{2},f_{3})$, $\mathbb{D}=(d/dt,d/dt,d/dt)$ and $\mathbf{U}=(U_{1},U_{2},U_{3})$ where
 $U_{1}(v_{1},v_{2},v_{3})=-v_{2}-v_{3}$, $U_{2}(v_{1},v_{2},v_{3})=v_{1} + \alpha v_{2}$ and $U_{3}(v_{1},v_{2},v_{3})=\beta + v_{3}(v_{1}-\gamma)$. For certain values of the parameters $\alpha,\beta,\rho$, the system is known to have chaotic solutions \citep{caotic-rosseler, caotic-rosseler2}.
 
Suppose now that we have access to a particular time series that was generated by this system. Our objective is to find some system of differential equations that can explain the behaviour of the measurements. The SINDy method works by choosing a large dictionary of functions and regressing discretisations of 
$
\frac{\partial f_1}{\partial t}, \dots, \frac{\partial f_H}{\partial t}
$
on the dictionary. The dictionary in question can be formed, for example, by collecting polynomial powers of $f_h$, $h=1,\dots,H$, spatial derivatives of $f_{1}\dots,f_{H}$ and trigonometric functions of $t$. A concrete simple example of such a dictionary in the case in which $H=1$ is the following:
$$
\mathcal{A}= \left\lbrace t,t^2, \sin(t), f_1, f_1^2, tf_1\right\rbrace.
$$
Of course in practice all derivatives are replaced by the corresponding finite differences taken from the measurements represented in $f$. %Further details on this point will be provided shortly. 
Having chosen a dictionary, we let $\mathcal{A}=(A_{1},\dots,A_{p})$ be the vector collecting all members of the dictionary.

Using the observations of the dynamical system, a regression model can be fitted to find the combination of the elements of the dictionary of functions that adequately explains the behavior of $\frac{\partial f_1}{\partial t}, \dots, \frac{\partial f_H}{\partial t}$. That is, we look for a vector of regression coefficients $\mathbf{c}=(c_1,\dots,c_p)$ such that for all $h=1,\dots, H$
\begin{equation}
    \frac{\partial f_{h}(t)}{\partial t} \approx \sum\limits_{i=1}^{p} c_{i,h}. A_i(t).
    \label{eq:regression}
\end{equation}

The regression model has to be learned using the available data. This regression problem could be solved in principle using least-squares. However, the ordinary least-squares regression estimator is ill-defined in cases in which the number of predictor variables $p$ is larger than the number of observations. Since the analyst is usually uncertain about the number of elements in the dictionary needed to adequately model the system of interest, the method used to solve the regression problem at hand should allow for large number of predictor variables (possibly larger than the number of observations) and automatically estimate sparse models, that is, generate accurate models that only use a relatively small fraction of predictor variables. The Lasso regression technique  is perfectly suited for this task. The Lasso is an $\ell_{1}$-regularised least-squares regression estimator, defined as follows. For $h=1,\dots,H$ such that $f_{h}$ is observable we let
\begin{equation}
    \mathbf{c}^{*}_{h} = \arg\min_{\mathbf{c}_{h}\in\mathbb{R}^{p}}  \sum\limits_{l}\left(\frac{\partial^n f_{h}}{\partial t^n}(t_l) - \sum\limits_{i=1}^{p} c_{i,h} A_i(t_l)\right )^2 + \lambda ||\mathbf{c}_{h}||_1 , 
    \label{eq:LASSO}
\end{equation}
where $\lambda >0$ is a tuning constant, measuring the amount of regularization. It can be shown (\citep{sparse-book}) that the $\ell_{1}$ penalty encourages sparse solutions and that the larger $\lambda >0$ the sparser the solution vector  $\mathbf{c}^{*}_{h}$ will be. In practice, $\lambda$ is usually chosen by cross-validation. 
Note that any other sparse regression technique could have been used to estimate the coefficients. We prefer the Lasso due to its simplicity and the wide availability of efficient algorithms to compute it. See for example \citep{CD}.

The main assumption behind this methodology is that the dynamical system that generated the data at hand can, in reality, be at least approximated using a sparse model, that is, that the vectors $\mathbf{c}^{*}_h$ in  \eqref{eq:LASSO} are either exactly or approximately sparse. This hypothesis is known to hold for several dynamical systems of interest in different fields of science. See \citep{sindy}. If the hypothesis holds, we can expect the Lasso estimates to select only a few elements of the dictionary, namely, those that do a good job at explaining variations in the response variable (\citep{sparse-book}).

In \citep{DataDrivenPDE2}, \citep{DataDrivenPDE} and \citep{sindy} the authors propose to use an ad-hoc linear regression estimator based on iteratively thresholding the least-squares estimator and applied this method only to first order systems. Through extensive numerical experiments, they show that this methodology is able to learn systems of partial differential equations that adequately model the dynamical system that generated the data. Unfortunately, if some variables are latent, that is, if one is unable to measure at least one of $f_{1},\dots, f_{H}$, the approach described above cannot be used directly. Next, we describe a way of extending this methodology to deal with the case in which some variables are latent.

\subsection{Our proposal: L-ODEfind}
\label{sec:methods_our_proposal}

To accommodate the possibility of latent variables we note that, for a large class of dynamical systems, it is possible to reconstruct a trajectory (equivalent to the original one) given only one of the model variables, using its higher order derivatives (\citep{Takens}). Moreover, we recall that in the case of a linear system of $n$ ordinary differential equations there is an equivalence between this multidimensional system and a single differential equation of order $n$, which we can interpret as latently including the information of the other $n-1$ unobserved variables. \citep{GPoMo} also makes use of higher order time derivatives to deal with unobserved variables.

Based on this ideas, we propose to augment the methodology developed in \citep{DataDrivenPDE2}, \citep{DataDrivenPDE} and \citep{sindy} by choosing the target variable to be a higher-order time derivative, to tackle situations in which not all relevant variables are observed. 
We estimate the coefficients of the dynamical system using the Lasso estimator \eqref{eq:LASSO}. As mentioned earlier, we chose to use the Lasso due to its simplicity, the abundance of theoretical guarantees on its performance (\citep{sparse-book}) and the availability of efficient algorithms to solve the convex optimization problem that defines the estimator. The choice of the tuning constant $\lambda$ in \eqref{eq:LASSO} is done by 10-fold cross validation, using the LassoCV method from sklearn \citep{sklearn} with a maximum number of steps equal to 10000, and 100 candidate $\lambda$s.
After estimating the regression coefficients, we build a forecasting method by integrating the retrieved differential equation.
We call this method Latent ODE find (L-ODEfind).

For instance, suppose we have observed a single time series $f$ ($H=1$). If we choose as a target variable the third time derivative and we use polynomial combinations up to degree 2 of the series and derivatives up to order 2 as regressors, equation \eqref{eq:regression} becomes
\begin{equation}
    \frac{\partial^3 f(t)}{\partial t^3} \approx \sum\limits_{i=1}^{p} c_{i} A_i(t),
    \label{eq:target3poly2}
\end{equation}
and    
$$
\mathcal{A} = \left\lbrace 1, f, \frac{\partial f(t)}{\partial t}, \frac{\partial^2 f(t)}{\partial t^2},
    f^2, \left(\frac{\partial f(t)}{\partial t}\right)^2, \left(\frac{\partial^2 f(t)}{\partial t^2}\right)^2, 
    f\frac{\partial f(t)}{\partial t}, 
    f\frac{\partial^2 f(t)}{\partial t^2},
    \frac{\partial f(t)}{\partial t}\frac{\partial^2 f(t)}{\partial t^2}
    \right\rbrace.
$$
This regression model is then fitted using the Lasso \eqref{eq:LASSO}, as described earlier.

\subsection{Evaluation with predictions}
\label{sec:methods_smape}

Evaluating the equations found by these methods can be challenging. The evaluation of a method that aims at recovering the differential equation behind the observed data needs to be done in different ways depending on the information available.

% When there are hidden variables, we propose to evaluate the solutions found by a given method for learning differential equations by measuring their forecasting accuracy. That is, the differential equations found by a method are integrated to obtain predictions which can then be compared with the true values. 

% The generating differential equation coefficients might or might not be available and all the variables involved might or might not be observed.

In the case of a simulation where the coefficients of the differential equation are known and the variables are fully observed, the adjusted coefficients and the true ones can be compared using the mean squared error.  If not all variables are observed, the adjusted coefficients refer to a different differential equation that in most cases cannot be obtained analytically (exceptions are, for instance, linear ODEs). In this case, the coefficients comparison cannot be done and another way of evaluating the method is needed.  The same happens when working with real world data where the differential equation behind is not known. In order to evaluate a method in this context, we propose to integrate the fitted differential equation to make predictions for different time horizons and compare them to the observed data by using the symmetric mean absolute percentage error (SMAPE). Using a time horizon of $n$, the SMAPE is defined as:

$$SMAPE(A,F) = \frac{1}{n}\sum_{t=1}^n \frac{|F_t- A_t|}{(|A_t|+|F_t|)/2}, $$
where $A_t$ is the real value and $F_t$ the forecasted value. 
% The predictions  are a way of evaluating the adjusted model and the differential equation found by the method when the target differential equation is not known. 
If the predictions made by method A are better (lower values of SMAPE) than the ones made by method B, the differential equation found by the method A is deemed better than the one found by method B. 

\subsection{Implementation details}

All the experiments in this paper were performed using Python 3.8, except for GPoMo , which was performed using R 3.6 \citep{R} and the GPoM package \citep{gpomopackage}. For L-ODEfind we use our own Python implementation (available in \url{https://github.com/agussomacal/L-ODEfind}). The integration of the differential equations was done using odeint from python Scipy library \citep{scipy}.

%% file: sections/Results.tex
In this section we first compare L-ODEfind  with GPoMo for the problem of learning an ordinary differential equation with no hidden variables. Then, we compare the performance of these methods in simulated systems with hidden variables and finally in temperatures series provided by RTE.

\subsection{Names abbreviation}
The names abbreviation used in this section can be found in Table \ref{table:model-names}.
Target time derivative is the degree of the fitted differential equation and poly degree is the maximum degree of the polynomial combinations of the derivatives used as regressors. An example of the regression problem with target derivative 3 and poly degree 2 can be found in equation \eqref{eq:target3poly2}.   Notice that the number in the model name refers to the degree of the differential equation. 
\begin{table}[H]
\centering
\begin{tabular}{|l|l|l|}
\hline
\textbf{Method} & \textbf{Parameters}               & \textbf{Name} \\ \hline
L-ODEfind         & Target derivative: 1, poly degree: 3 & L-odefind1      \\ \hline
L-ODEfind         & Target derivative: 2, poly degree: 3 & L-odefind2      \\ \hline
L-ODEfind         & Target derivative: 3, poly degree: 3 & L-odefind3      \\ \hline
GPoMo           & Target derivative: 2, poly degree: 3 & GPoMo2        \\ \hline
GPoMo           & Target derivative: 3, poly degree: 3 & GPoMo3        \\ \hline
\end{tabular}
\caption{Models names abbreviations used in the graphics above.}
\label{table:model-names}
\end{table}

% For instance, odefind1, adjust the coefficients of the equation \ref{target1} where the target derivative is 1 (the degree of the differential equation is 1) and the polynomial only considers the variable x. Equation \ref{target2} refers to odefind2, where the target derivative is 2 and equation \ref{target3} refers to odefind3, where the target derivative is 3. 
% \begin{align}
% \frac{dx}{dt} &= poly(x) \label{target1}\\
% \frac{d^2x}{dt^2} &= poly\left(x, \frac{dx}{dt}\right)  \label{target2}\\
% \frac{d^3x}{dt^3} &= poly\left(x, \frac{dx}{dt}, \frac{d^2x}{dt^2}\right) \label{target3}
% \end{align}

%------------------------------------------------
\subsection{Simulated data with fully observed variables}
\label{sec:fullinfo}

We compare the performance of L-ODEfind with that of GPoMo for the task of learning an ordinary differential equation with no hidden variables. Note that since in this case there are no hidden variables, our method coincides with SINDy. The goal of this comparison is to highlight the fact that, because L-ODEfind solves a continuous optimization problem and GPoMo approximately solves a combinatorial optimization problem, L-ODEfind can be orders of magnitude faster that GPoMo.

We generated data using the Lorenz attractor equations:

 \begin{align}
\frac{dx}{dt} &= \sigma (y-x), \nonumber \\
\frac{dy}{dt} &= x (\rho-z) - y, 
\label{eq:lorenz} \\
\frac{dz}{dt} &= xy- \beta z, \nonumber 
\end{align}
using the coefficients $\sigma = 10$, $\rho= 28$ and $\beta =  \frac{8}{3}$, for twenty different random starting conditions, with distribution $N(0,1)$. The differential equation was integrated using a discretization step of 0.01.
Then, we fitted the resulting datasets using GPoMo and L-ODEfind. For GPoMo we set a maximum number of steps of 10240. 
We compare the true coefficients of the differential equations with the ones found by each model using as a metric the mean square error (MSE). All the coefficients were taken into account for computing the MSE, including the ones that are supposed to be zero. We average the MSE corresponding to different random starting conditions and report this as our goodness of fit metric. 

\begin{figure}[H]
    \includegraphics[width=0.9\textwidth]{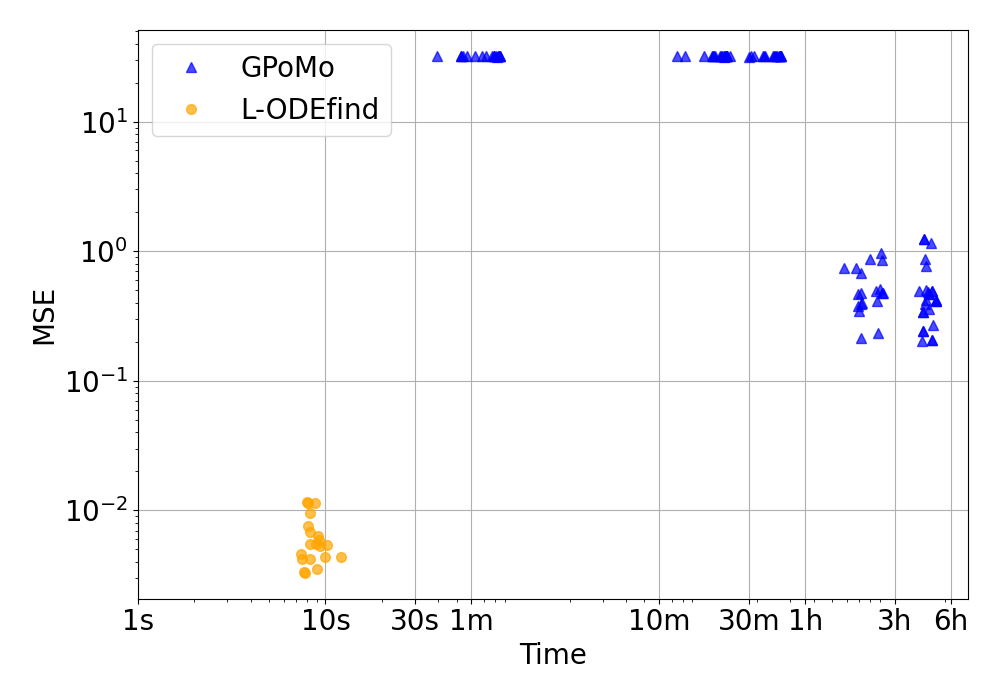}
    \caption{Comparison between L-ODEfind and GPoMo when finding the coefficients of data generated by the Lorenz attractor differential equations. We plot the average MSE across different random initial conditions for the system, and the time (in seconds) needed by the algorithm to fit the model. }
    \label{fig:mse_lorenz}
\end{figure}

Figure \ref{fig:mse_lorenz} shows that L-ODEfind method is nearly two orders of magnitude more accurate than the best GPoMo case. On the other hand, L-ODEfind is also orders of magnitude faster, taking less than 10 seconds to compute accurate approximations of the true coefficients while GPoMo takes hours. 

%------------------------------------------------
\subsection{Simulated data with hidden variables}
\label{sec:latent}

In this section we use three ODE systems, an oscillator, and the Rossler and Lorenz systems, as examples to evaluate and compare, using the methodology explained in \ref{sec:methods_smape}, the accuracy of several L-ODEfind and GPoMo models. For a given differential equation, we generated time series of length $5000$ points and integration step of $dt=0.01$ for $20$ different initial conditions. Following the methodology described in \ref{sec:methods_smape} we fit each series with GPoMo or L-ODEfind and then integrate the equation found in each case to predict the values of the series for several time horizons and compute the corresponding SMAPE. The number of maximum iterations for GPoMo is set to 5120.

%------------------------------------------------
\subsubsection{Example 1: Oscillator}

We start by considering an oscillator, which is a first order linear system of two variables. An equation of order two involving only one of the variables can be derived. So, in this case, the problem of having hidden variables can be effectively tackled by choosing a higher order time derivative (order two) as the target.

We used the two variables oscillator equation which can be written in  general form as:

% {'a': 0.1, 'b': -1, 'c': 1, 'd':0}

\begin{equation}
\frac{d}{d t}
    \begin{pmatrix}
  x\\ 
  y
\end{pmatrix} = \begin{pmatrix}
  a & b\\ 
  c & d
\end{pmatrix} \cdot \begin{pmatrix}
  x\\ 
  y
\end{pmatrix}
\end{equation}
where $x$ and $y$ are the variables and $a,b,c,d$ are the coefficients of the linear equation that links the variable with its derivatives. If we only had access to the variable $x$, this system could be rewritten in a second order differential equation taking the form:

\begin{equation}
\frac{d^2 x}{d t^2} \quad = \quad (a+d) \frac{dx}{dt} - (ad-bc)x \quad = \quad \beta \frac{dx}{dt} + \alpha x.
\label{eq:osc}
\end{equation}
 
\begin{figure}[H]
\centerline{
\subcaptionbox{Mean SMAPE versus time horizon.}{\includegraphics[width=0.6\textwidth]{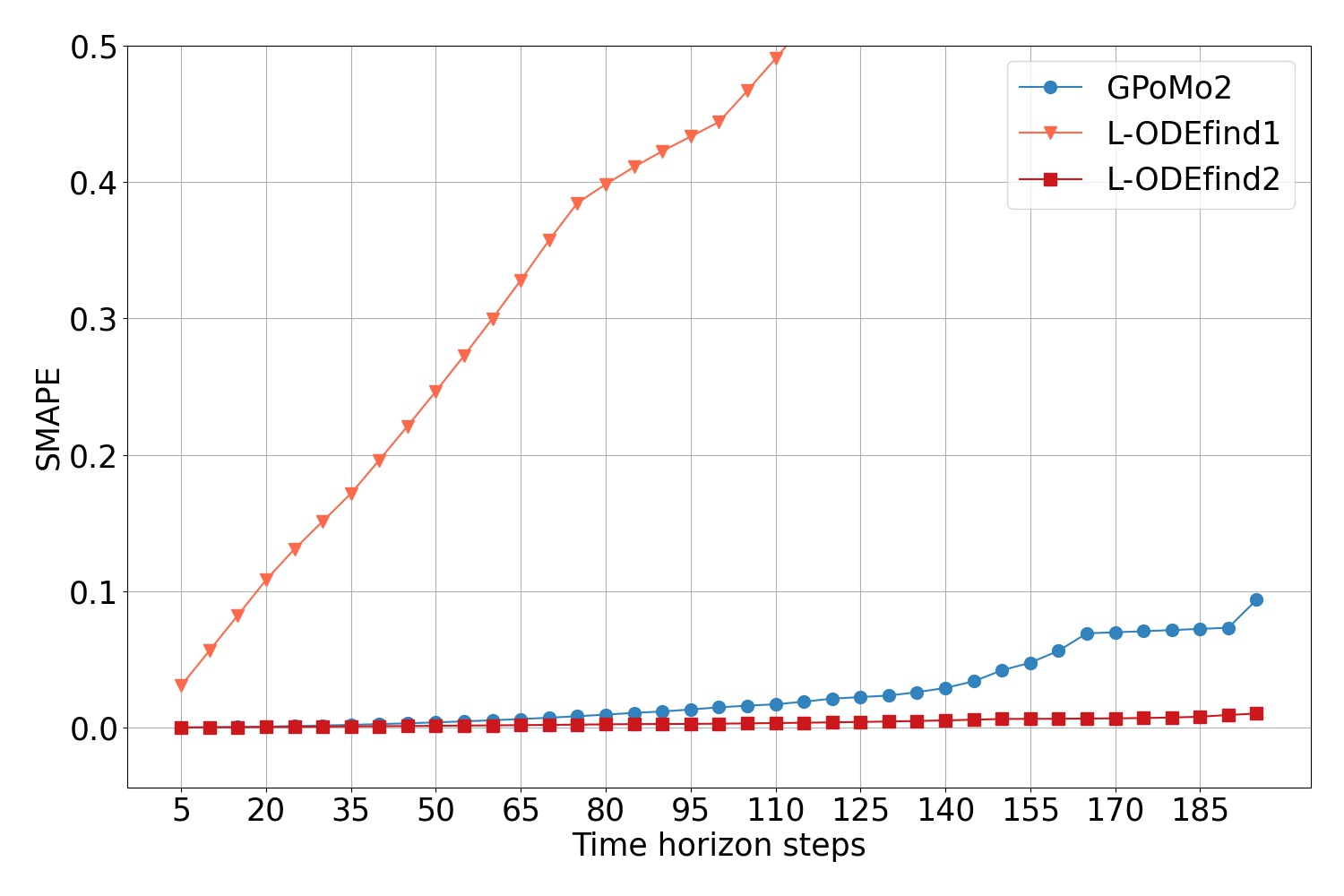}}
\label{fig:smape_osc}
\subcaptionbox{Fitting time. }{\includegraphics[width=0.4\textwidth]{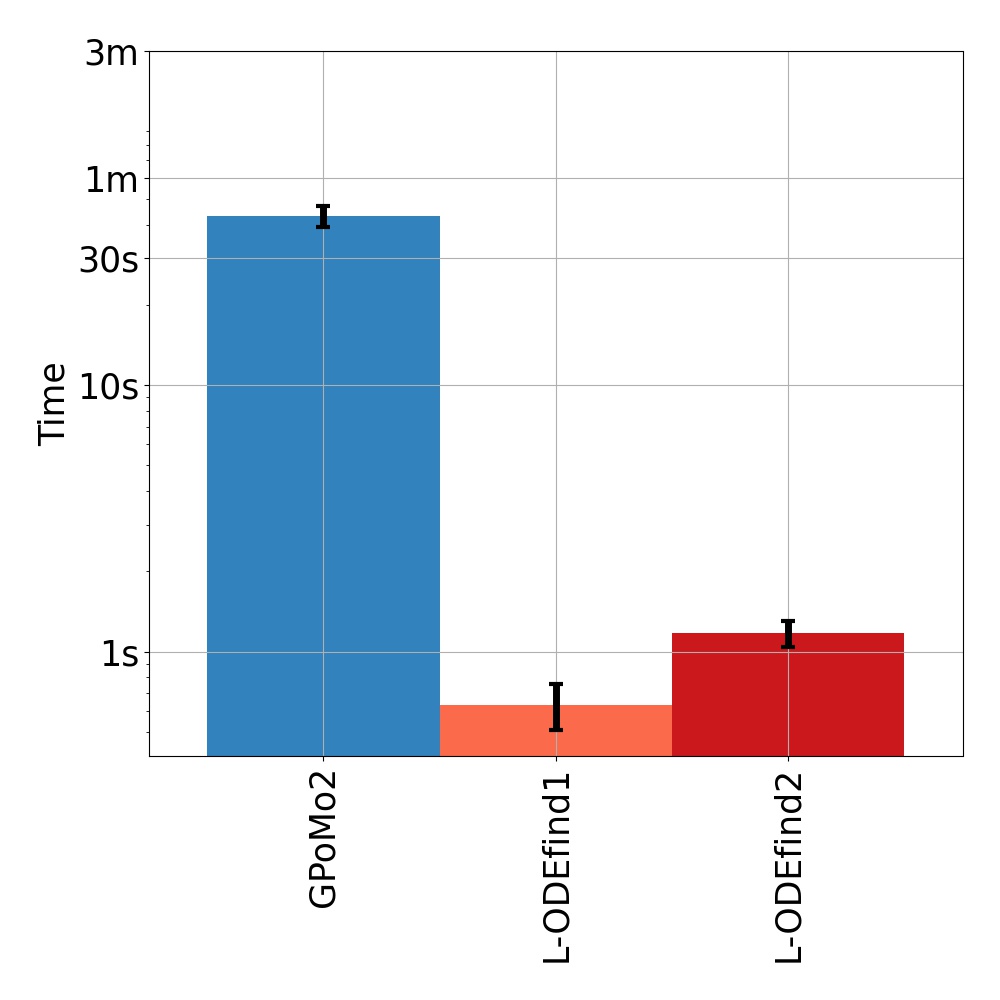}}
\label{fig:times_osc}
}
\caption{Prediction accuracy and fitting time for L-ODEfind and GPoMo models when data comes from the oscillator system with x as observed variable (y hidden). }
\label{fig:osc}

\end{figure}

In our experiments we set $a=0.1,b=-1,c=1,d=0 \: (\alpha=-1, \beta=0.1)$ and we only observed the variable $x$ so the corresponding second order equation derived as in \ref{eq:osc} is $0.1\frac{dx}{dt} - x$.
We can see in Figure \ref{fig:osc} (a) that using the second derivative in time as target gives the lowest prediction error for both GPoMo and L-ODEfind, as expected from equation \eqref{eq:osc}. We also see that the model found by L-ODEfind manages to approach the true model much better than GPoMo as it maintains an SMAPE below $0.02$ in horizons were GPoMo has already arrived to $0.1$. Moreover, when looking at the coefficients found by GPoMo ($\alpha=-0.99506\pm0.00372$ and $\beta=0\pm0$) and L-ODEfind ($\alpha=-0.99965\pm 0.00030$ and $\beta=0.09963\pm0.00013$) are within a small tolerance the expected from equation \eqref{eq:osc}. 
Finally, when looking at the fitting time (Fig. \ref{fig:osc} (b)), we find that L-ODEfind is around $50$ times faster than GPoMo.

%------------------------------------------------
\subsubsection{Example 2: Rossler}

Next, we consider a more complicated case, the Rossler system, which is a non-linear (quadratic) system:
 \begin{align}
\frac{dx}{dt} &= -x-z, \nonumber \\
\frac{dy}{dt} &= x + ay,  \\
\frac{dz}{dt} &= b+z(x-c). \nonumber 
\label{eq:rosseler}
\end{align}
If only variable $y$ is observed ($x$ and $z$ hidden) an equation of order $3$ and polynomial degree $2$ can be deduced for $y$ \citep{rossler-deduction} .

% so we can still compare the results using both the MSE of the estimated coefficients \eze{pero reportamos el MSE de los coeficientes?} and the SMAPE.

In our experiments, we set  $a=0.52, b=2, c=4$ and only observe $y$ ($x$ and $z$ latent). L-ODEfind is consistently faster than GPoMo although, in this example, GPoMo3 and L-ODEfind3 have almost the same prediction SMAPE. Notice that the SMAPE is less than 0.2 for all models with target time derivative 2 or 3 up to 125 time horizon steps.

\begin{figure}[H]
\captionsetup{width=1.2\textwidth}
\centerline{
\subcaptionbox{Mean SMAPE versus time horizon.}{\includegraphics[width=0.6\textwidth]{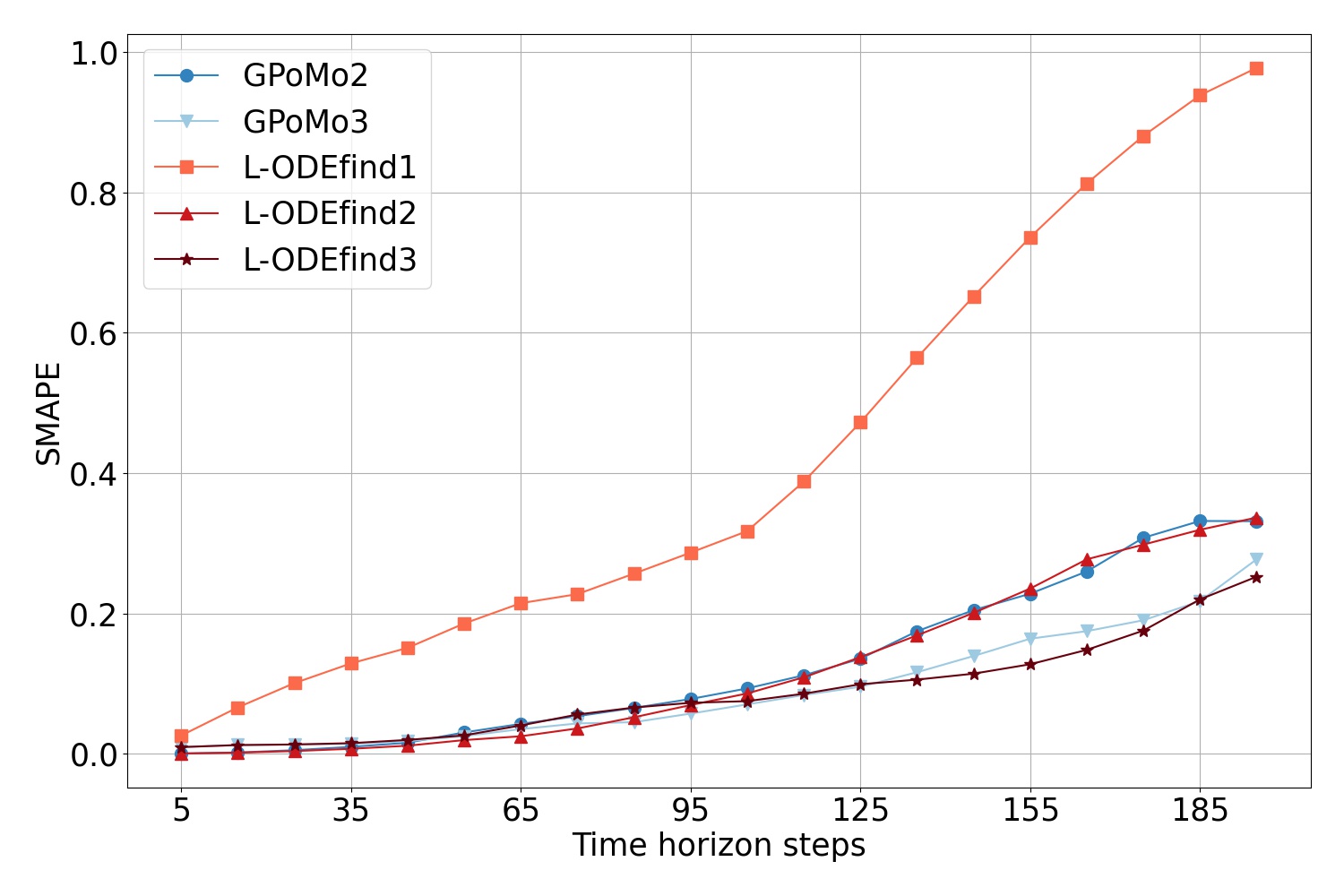}}
\subcaptionbox{Fitting time. }{\includegraphics[width=0.4\textwidth]{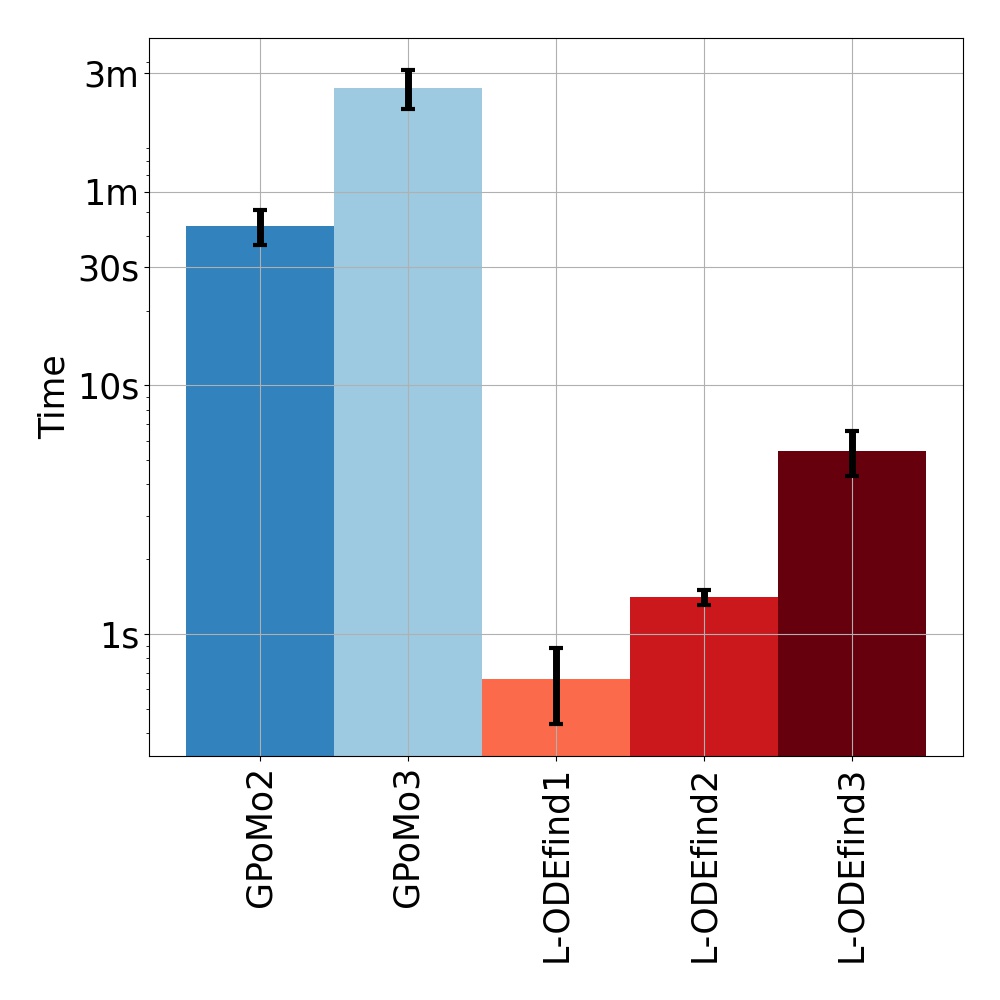}}
}
\caption{Prediction accuracy and fitting time for L-ODEfind and GPoMo models when data comes from the Rossler system with y as observed variable (x and z hidden).}
\label{fig:rosseler}
\end{figure}

Therefore, when considering systems where an analytical solution can be deduced, such us the oscillator ($x$ observed, $y$ hidden) and Rossler system ($y$ observed, $x$ and $z$ hidden), both methods perform very well in terms of SMAPE prediction. 

\begin{figure}[H]
 \captionsetup{width=1.2\textwidth}
\centerline{
\subcaptionbox{Mean SMAPE versus time horizon.}{\includegraphics[width=0.6\textwidth]{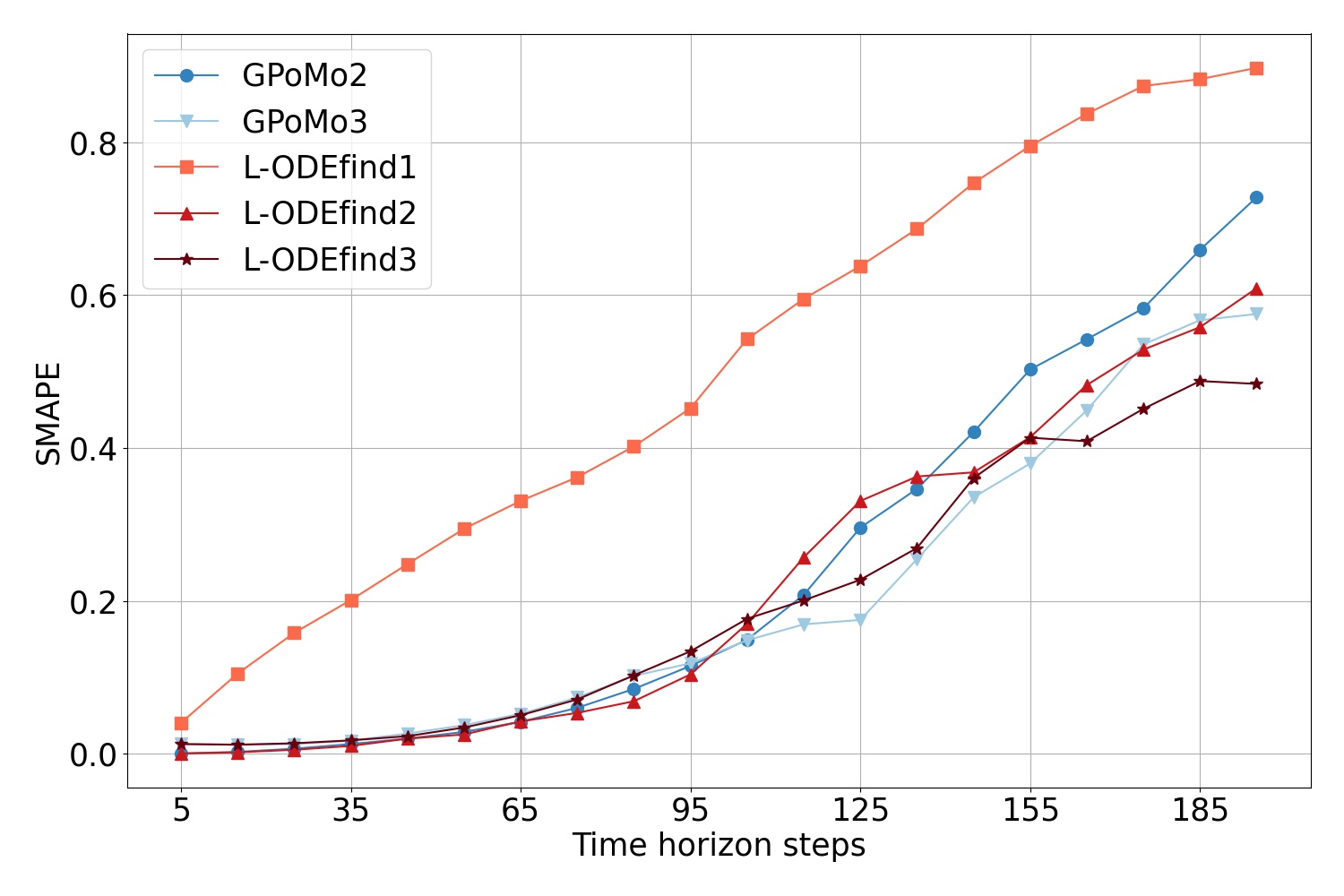}}
\subcaptionbox{Fitting time.}{\includegraphics[width=0.4\textwidth]{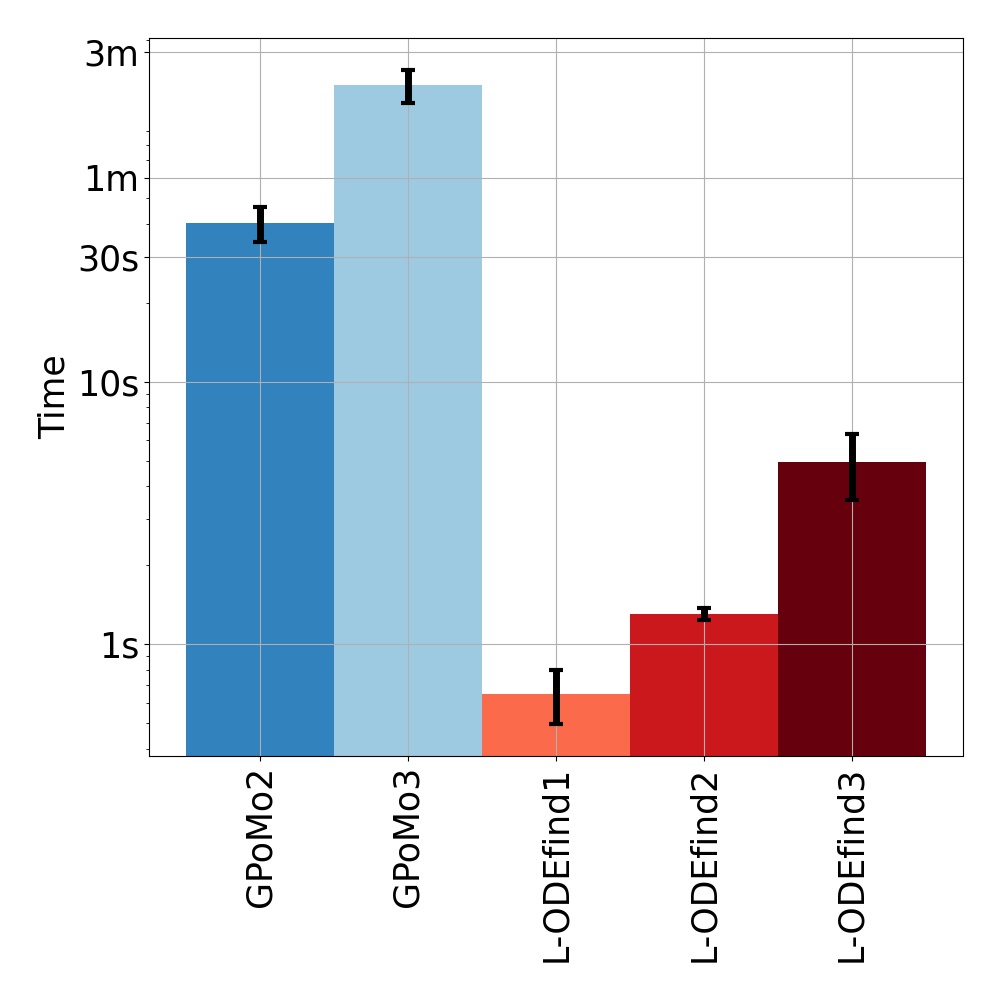}}
}
\caption{Prediction accuracy and fitting time for L-ODEfind and GPoMo models when data comes from the Rossler system with x as observed variable (y and z hidden).}
\label{fig:rosseler-x}
\end{figure}

Next, we consider the Rossler system in the case of variable $x$ observed ($y$ and $z$ hidden). In this scenario, there is no analytical solution using only polynomials \citep{rossler-deduction}. Interestingly, we can see that using higher order time derivatives as target, both GPoMo and L-ODEfind find a differential equation that is able to provide predictions whose accuracy is comparable (for short-term horizons) to the previous case ($y$ observed) where there was an analytical solution (Figure \ref{fig:rosseler-x}). The fitting times continue to show that L-ODEfind is consistently faster than GPoMo.

%------------------------------------------------
\subsubsection{Example 3: Lorenz attractor}

Here we consider the Lorenz system \citep{lorenz1963}, discussed in Section \ref{sec:fullinfo}, where no differential equation using only polynomials can be derived for $x$ as the only observed variable. 

\begin{figure}[H]
 \captionsetup{width=1.2\textwidth}
\centerline{
\subcaptionbox{Mean SMAPE versus time horizon.}{\includegraphics[width=0.6\textwidth]{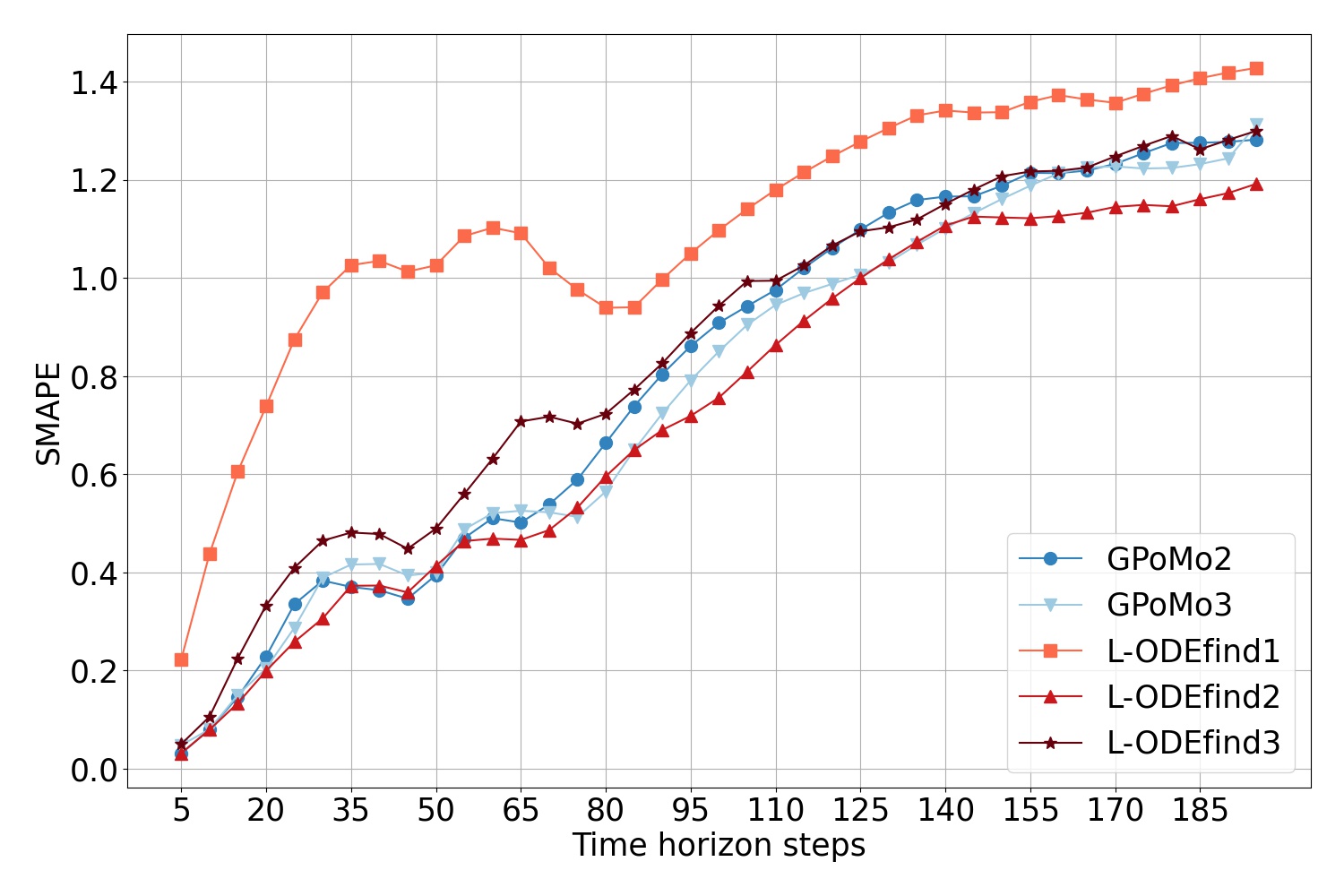}}
\subcaptionbox{Fitting time.}{\includegraphics[width=0.4\textwidth]{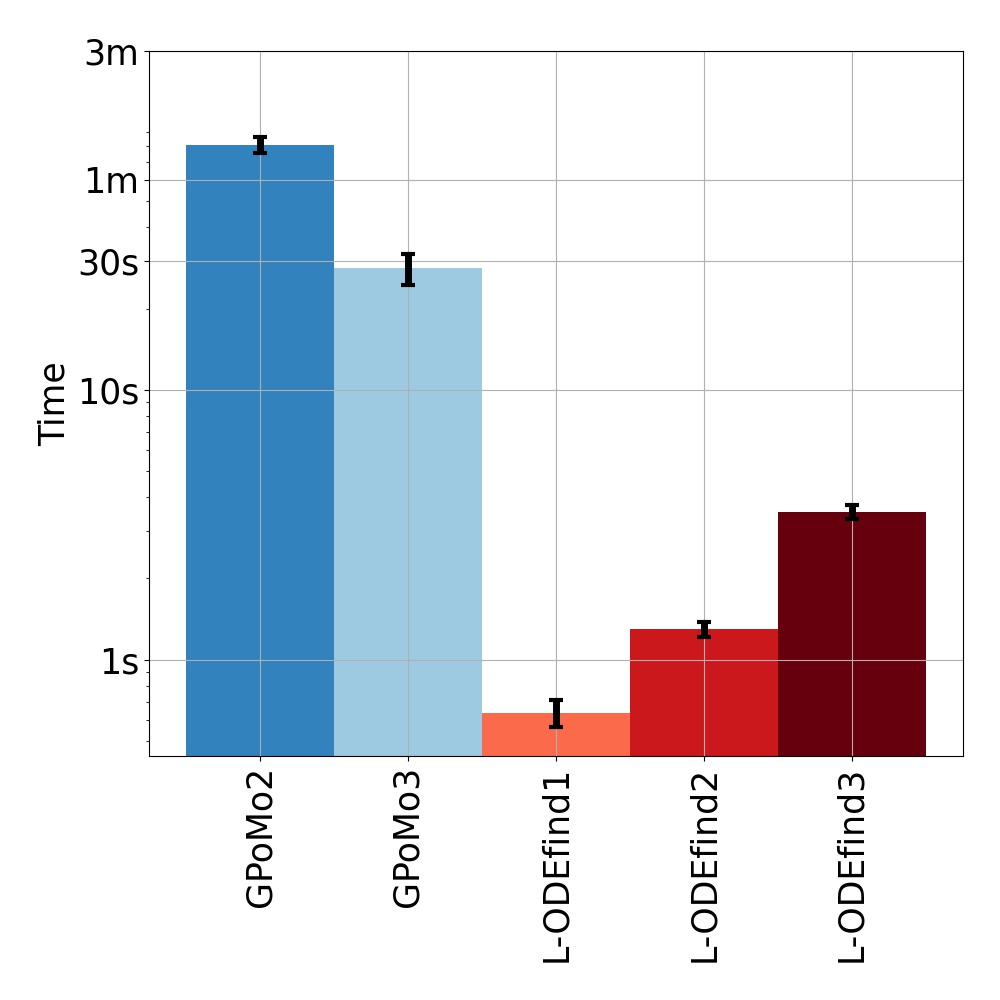}}
}
\caption{Prediction accuracy and fitting time for L-ODEfind and GPoMo models when data comes from the Lorenz system with x as observed variable (y and z hidden).}
\label{fig:lorenz}
\end{figure}

Following the same methodology as before, we tried different target derivatives for both GPoMo and L-ODEfind to fit observed variable $x$ ($y$ and $z$ served). We found that using higher order time derivatives (in particular second order) helps to find models that can approximate better the observed time series when integrated, although the prediction accuracy suffers from the added complexity of the problem (Fig. \ref{fig:lorenz}). As a consequence, the accuracy degrades faster reaching an $SMAPE=0.5$ as soon as $35$ time steps while in the previous cases this was attained around $185$ steps. For a narrow difference again L-ODEfind outperformed GPoMo while also keeping fitting times $4$ to $40$ times faster than GPoMo.

%------------------------------------------------
\subsection{Temperature series provided by RTE}
\label{sec:rte}

In this section, we fit different models to temperatures series provided by the Réseau de Transport d'Électricité (RTE).
The data consist of 200 hourly measured temperatures time series. These time series correspond to temperatures in Paris along a year for 200 different possible years or scenarios. The time series are not measured temperatures nor the output of a simulation, but rather the result of a reanalysis process. Also, some relevant variables for modeling the atmospheric system are not available to us, for instance, wind and pressure.
Our temperature time-series are not historical measurements but can be thought as a possible realization of the temperature in Paris. These temperature time series have $365\times24 = 8760$ time points. 

In order to evaluate the methods, we select $39$ out of the $200$ temperature time series (due to  GPoMo's computation time),  fit the different models and use them to predict for short time horizons. The fit was done with the first $8560$ time points and the prediction was evaluated with the following time points (up to a time horizon of $15$ time points). 

In tackling this complex problem we want to analyze the performance of \textbf{L-ODEfind} and \textbf{GPoMo} in the forecasting task and compare their behavior to classical forecasting methods \textbf{naive predictor} (for every time horizon, predicts the average of the last 24 time points) and \textbf{exponential smoothing (ES)} (triple exponential smoothing with an additive seasonal component\footnote{We used the implementation available in the Python library sktime \citep{sktime}}). 

\begin{figure}[H]
 \captionsetup{width=1.2\textwidth}
\centerline{
\subcaptionbox{SMAPE vs time horizons in hours for different predictive methods.}{\includegraphics[width=0.6\textwidth]{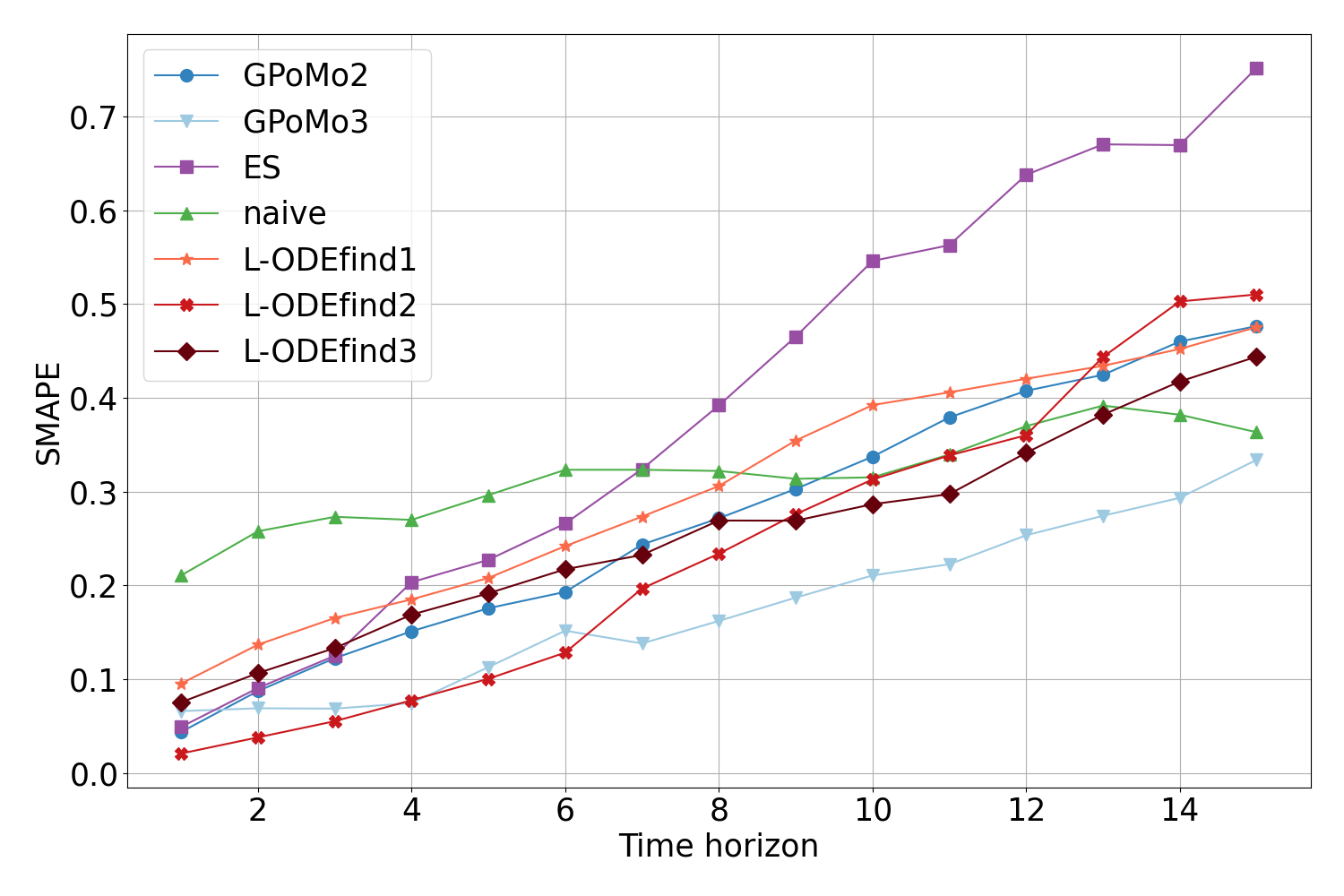}}
\subcaptionbox{Time to fit the different models in log scale. }{\includegraphics[width=0.4\textwidth]{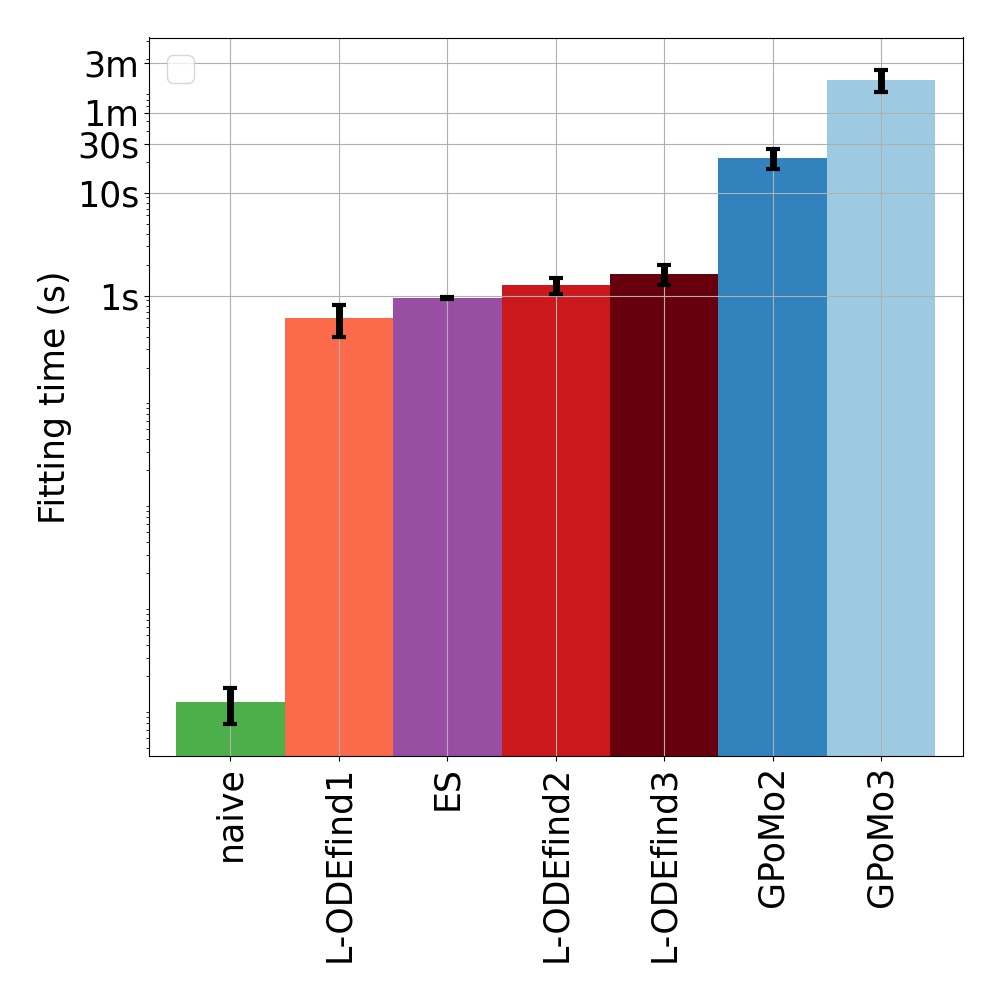}}
}
\caption{SMAPE and fitting times for different models applied on a real world data set of temperatures. }
\label{fig:rte15}
\end{figure}

In Figure \ref{fig:rte15} the SMAPE of the prediction for different time horizons in hours is displayed. L-ODEfind2 gives the lowest SMAPE for time horizons lower or equal to $6$ hours, whereas GPoMo3 is better for time horizons greater than $6$ hours. Notice that for time horizons greater than $6$ hours, the naive predictor performs better than exponential smoothing, giving a rough idea of the reasonable predictability horizon that forecasting methods can give. In any case, both L-ODEfind and GPoMo give better forecasts than both ES and naive when the target derivative is $3$.  
As can be seen in figure \ref{fig:rte15} (b), the fitting times for GPoMo are much greater than L-ODEfind, whereas L-ODEfind and exponential smoothing have similar fitting times.

%% file: sections/Conclusions.tex
In this paper we addressed the problem of recovering differential equations from data where not all variables are observed by enhancing the approach outlined in \citep{DataDrivenPDE2}. We tested this approach in simple and complex ODE systems and consistently found that the proposed approach of using time derivatives of higher order as target regressing variables allows to find models whose future predictions are more reliable than only using first order derivatives.
We also compared our method to GPoMo and found that: (i) our proposal is orders of magnitude faster that GPoMo, and (ii) our proposal learns models with comparable or even higher prediction accuracy for several dynamical systems.
Finally, we faced the challenge of addressing a real world problem and found that both L-ODEfind and GPoMo used as forecasting methods gave comparable results while at the same time outperforming classical forecasting methods, exponential smoothing and naive.
In summary, L-ODEfind proved to be an accurate and fast method for recovering ordinary differential equations from data with hidden variables. %Moreover, L-ODEfind can tackle partial differential equations scenarios, which is known to be a very challenging task.

%As future directions of work, an interesting problem is the extension of the method proposed in this paper to the problem of learning partial differential equations with latent variables. 
%It should be noted that, as GPoMo is specifically tailored for solving dynamical systems, it can't tackle partial differential equation scenarios whereas the L-ODEfind methodology is potentially able to handle partial differential equations. 

% In this work we left aside the exploration of situations modeled by partial differential equations where not all the information is available. However, in this direction there are some preliminary results showing that the proposed methodology can also be applied in this settings.

% Nowadays, collecting data is becoming popular along all disciplines. Methods that allow to harness this growing amount of data and give insights about the generative process behind, can be very useful. 